\documentclass[]{bytedance_seed}

\usepackage[toc,page,header]{appendix}
\usepackage{amssymb}
\usepackage{colortbl}
\usepackage{xcolor}
\usepackage[utf8]{inputenc}
\usepackage{enumitem} %

\usepackage{minitoc}

\title{Hyper-Bagel: A Unified Acceleration Framework for Multimodal Understanding and Generation
}

\author[*]{Yanzuo Lu}
\author[*]{Xin Xia}
\author[*]{Manlin Zhang}
\author{Huafeng Kuang}
\author{\\Jianbin Zheng}
\author{Yuxi Ren}
\author[\dagger]{Xuefeng Xiao}

\affiliation{ByteDance Seed}

\contribution[*]{Equal contribution}
\contribution[\dagger]{Corresponding author}

\abstract{
Unified multimodal models have recently attracted considerable attention for their remarkable abilities in jointly understanding and generating diverse content. 
However, as contexts integrate increasingly numerous interleaved multimodal tokens, the iterative processes of diffusion denoising and autoregressive decoding impose significant computational overhead. 
To address this, we propose Hyper-Bagel, a unified acceleration framework designed to simultaneously speed up both multimodal understanding and generation tasks. 
Our approach uses a divide-and-conquer strategy, employing speculative decoding for next-token prediction and a multi-stage distillation process for diffusion denoising. 
The framework delivers substantial performance gains, achieving over a 2x speedup in multimodal understanding. 
For generative tasks, our resulting lossless 6-NFE model yields a 16.67x speedup in text-to-image generation and a 22x speedup in image editing, all while preserving the high-quality output of the original model. 
We further develop a highly efficient 1-NFE model that enables near real-time interactive editing and generation. 
By combining advanced adversarial distillation with human feedback learning, this model achieves ultimate cost-effectiveness and responsiveness, making complex multimodal interactions seamless and instantaneous.
}

\correspondence{\email{xiaoxuefeng.ailab@bytedance.com}}

\checkdata[Project Page]{\url{https://hyper-bagel.github.io/}}

\begin{document}
\maketitle


\section{Introduction}


\begin{figure}
    \centering
    \includegraphics[width=\linewidth]{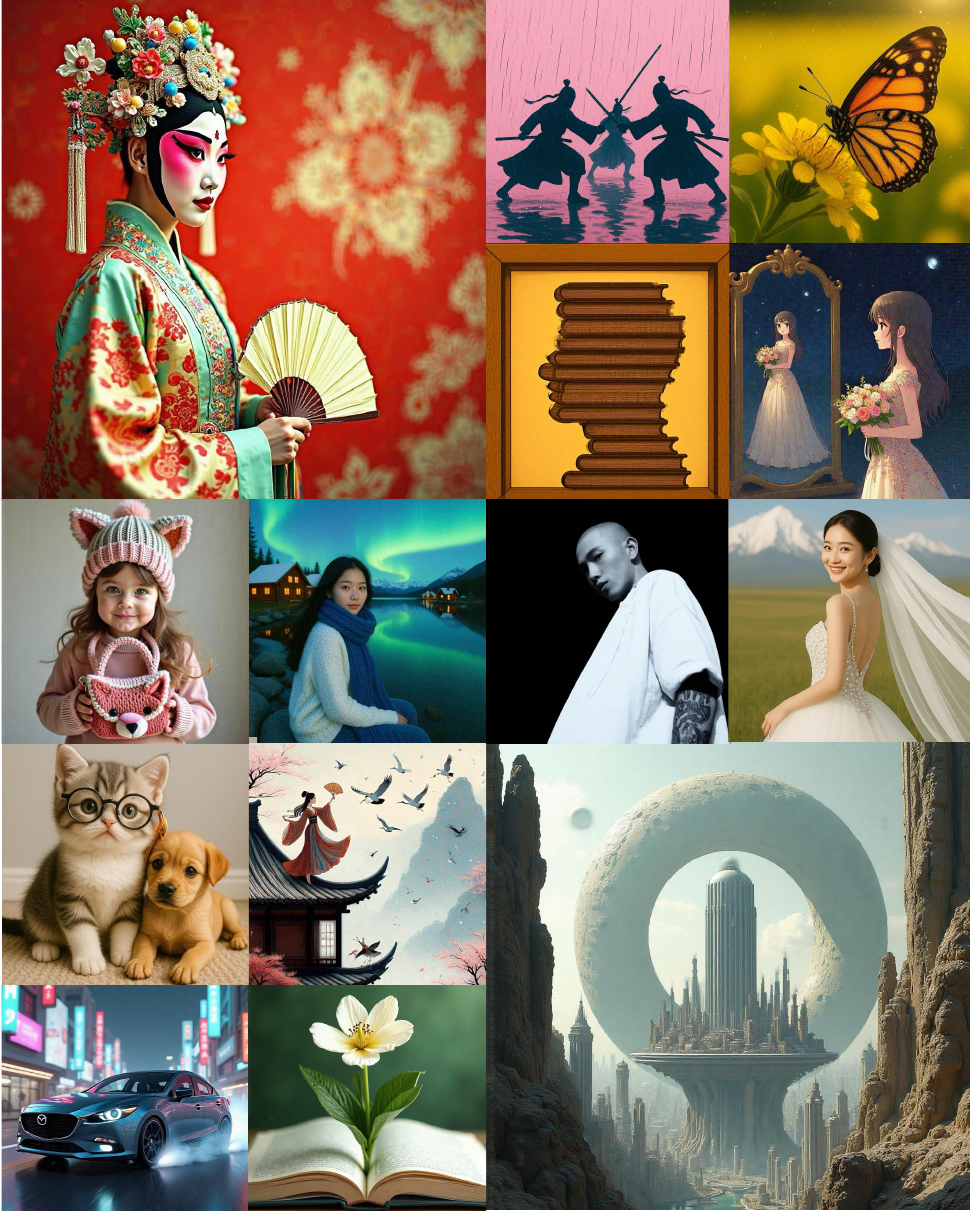}
    \caption{Image generation samples produced by our 6-NFE accelerated BAGEL model.}
    \label{fig:t2i_teaser}
\end{figure}

\begin{figure}
    \centering
    \includegraphics[width=0.9\linewidth]{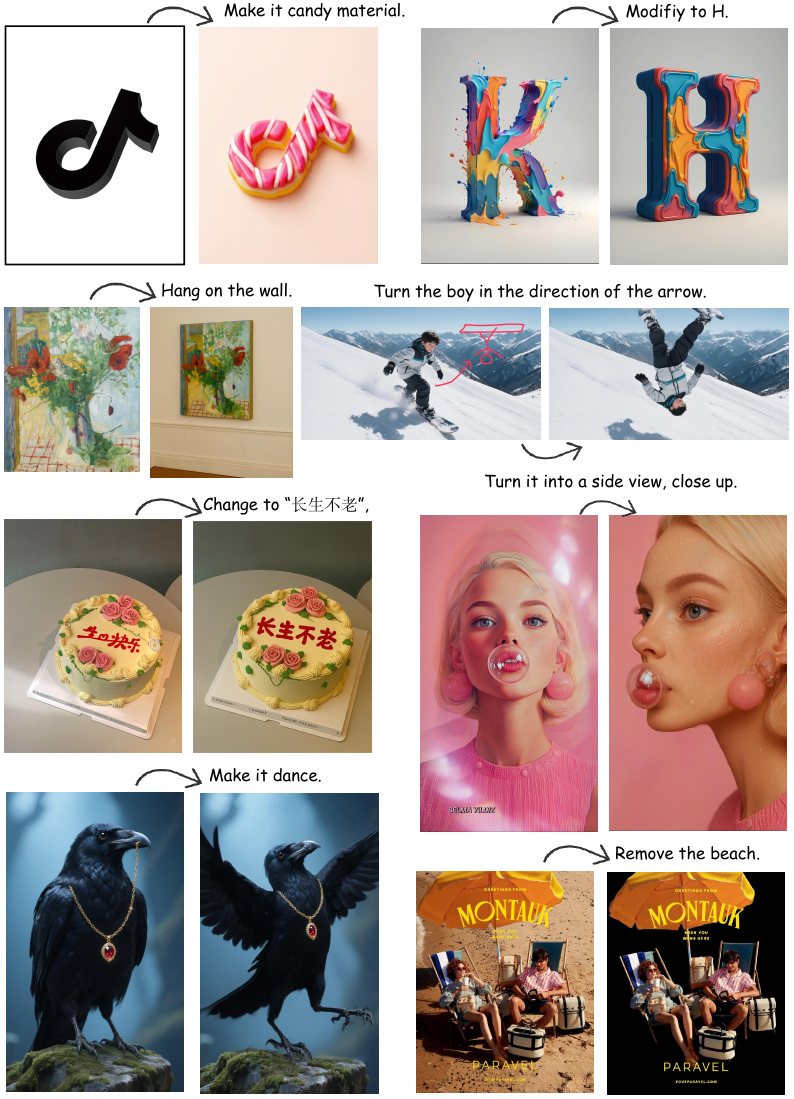}
    \caption{Image editing samples produced by our 6-NFE accelerated BAGEL model.}
    \label{fig:edit_teaser}
\end{figure}

The field of unified multimodal models has recently attracted considerable attention for their remarkable abilities in jointly understanding and generating diverse content, marking a promising leap forward in next-generation Generative Pre-trained Transformer (GPT) paradigms~\cite{zhou2024transfusion, team2024chameleon, sun2023emu, wu2025janus, xie2024show, deng2025emerging, achiam2023gpt, seawead2025seaweed, gao2025seedream, li2022next, liao2025mogao}.
While prior scaling approaches predominantly relied on standard image-text pairs, the innovative training paradigm incorporating interleaved sequences from BAGEL~\cite{deng2025emerging} has achieved notable successes, outperforming models specialized solely in understanding or generation across multiple benchmarks.
However, as contexts integrate increasingly numerous interleaved multimodal tokens, the iterative processes of diffusion denoising~\cite{ho2020denoising, lipman2022flow} and autoregressive decoding~\cite{brown2020language,achiam2023gpt} for next-token prediction impose significant computational overhead. 
To address this, a carefully engineered acceleration framework is crucial, one that delivers lossless speedups for both tasks simultaneously, thereby maintaining seamless knowledge transfer between understanding and generation modalities.

To realize this vision, we adopt a divide-and-conquer strategy and validate it on BAGEL.
For next-token prediction, we employ speculative decoding~\cite{leviathan2023fast, li2024eagle, cai2024medusa, li2025eagle3scalinginferenceacceleration} by training a lightweight draft model with fewer parameters and lower computational requirements. 
This draft model iteratively predicts multiple consecutive tokens, with the speculative output then batch-validated by the target model in parallel. 
This approach preserves the understanding capabilities of target model intact, converting its memory access bottleneck during autoregressive decoding into a computational one, resulting in over a 2x speedup. 
For diffusion denoising, we develop a multi-stage distillation process that decomposes objectives into three key dimensions to reduce sampling steps, i.e. Classifier-Free Guidance (CFG)~\cite{ho2022classifier} control, structural integrity, and image fidelity, each optimized with tailored algorithms. 
We successfully constrain sampling steps for text-to-image generation and image editing to within 6-NFE (Number of Function Evaluations), yielding speedups of 16.67x and 22x, respectively, while maintaining equivalent performance on GenEval~\cite{ghosh2023geneval} and GEdit-Bench~\cite{liu2025step1x} metrics compared to the baseline model.
See \cref{fig:t2i_teaser,fig:edit_teaser} for examples of image generation and editing via our 6-NFE model.

In adopting speculative decoding, we primarily adhere to the EAGLE-3~\cite{li2025eagle3scalinginferenceacceleration} training paradigm.
Our initial reproduction experiments reveal that a straightforward application of EAGLE-3 underperformed on BAGEL. 
Notably, community reproductions on the latest Vision-Language Model (VLM) Qwen3~\cite{yang2025qwen3} achieved only a 1.7x speedup in Tokens Per Second (TPS), diverging sharply from the 4-5x accelerations typically observed in Large Language Models (LLMs) such as Vicuna~\cite{zheng2023judging}, LLaMA~\cite{grattafiori2024llama}, and DeepSeek~\cite{liu2024deepseek}. 
We assume that this disparity stems from the complex interleaving of image and text tokens in multimodal sequences, where embedding spaces vary considerably. 
Therefore, the draft model's limited capacity fails to adequately model these variations, resulting in substantial prediction accuracy losses. 
Furthermore, BAGEL presents even greater challenges than standard VLMs because it must handle not only text and ViT image tokens but also the clean latent tokens prefilled after diffusion denoising.
To validate our hypothesis and resolve the issue, we design an efficient intermediate layer architecture that bridges the target and draft models, aggregating representations to facilitate the draft model's consecutive predictions more effectively. 
This led to a TPS improvement from a baseline of 98.3 to approximately 212.4 in the SGLang~\cite{zheng2024sglang} environment with chain decoding on a single A100 GPU, resulting in a 2.16x speedup.

Regarding diffusion distillation~\cite{ren2025hyper, lin2025diffusion,shao2025rayflow,lu2025adversarial,lin2024sdxl}, our primary objective is to preserve the model's full capabilities, including both control and quality perspectives. 
The capability in control refers to the ability to adjust the degree of instruction adherence and consistency maintenance with the original image in the editing scenario by text and image guidance scales, which is typically enabled through CFG.
Meanwhile, the quality of generated images can be assessed through two key dimensions, i.e. structural integrity and image fidelity, which are the most easily perceived factors during human preference evaluation.
Building on this foundation, we carried out distillation over three stages tailored for each capability.
The first stage focuses on CFG distillation, which aims to embed the text and image guidance scales alongside timesteps into a single forward pass as control conditions.
The latter two stages are similar to the previous Hyper-SD~\cite{ren2025hyper}, which involves adversarial distillation followed by score distillation, but several key improvements have been made that significantly contribute to the final performance.
Compared to the proposed TSCD in Hyper-SD, the most significant difference of our adversarial distillation lies in the design of a multi-head discriminator, which discriminates between fake and real latents at multiple scales to enhance structural integrity.
As for score distillation, to circumvent the drawback of overly smooth images generated by the Stochastic Differential Equation (SDE) based on the consistency sampler in DMD~\cite{yin2023one}, we propose a new method called DMDO.
This approach generates images via an Ordinary Differential Equation (ODE) employing the Euler discrete sampler, mirroring the baseline model's process.
A notable advantage of this method is that it eliminates the need for additional ODE-based~\cite{yin2023one,ren2025hyper} or GAN-based~\cite{yin2024improved,lin2024sdxl,ren2024byteedit} regularizers to counterbalance the KL divergence, unlike in DMD~\cite{yin2023one} or DMD2~\cite{yin2024improved}.

In line with advances like Hyper-SD~\cite{ren2025hyper}, we also developed a 1-NFE model for exceptional inference speed and cost-effectiveness, building upon our 6-NFE model which was distilled without compromising the performance and output quality of the original.
Specifically, this model is obtained through an extra adversarial process based on rectified flow, followed by human preference alignment using Reward Feedback Learning (ReFL)~\cite{xu2024imagereward}.
Inspired by the Adversarial Diffusion Pretraining (ADP) proposed by DMDX~\cite{lu2025adversarial}, we utilize an ODE-based objective for adversarial training in the style of rectified flow, which means that we will employ 6-NFE inference from pure noise at each iteration to sample several probability flows as training data.
Considering that BAGEL focuses more on knowledge transfer between modalities, we renew the reward model to the VLM-based HPSv3~\cite{ma2025hpsv3widespectrumhumanpreference} built upon Qwen2-VL~\cite{wang2024qwen2} that comprises much powerful visual understanding capability than the aesthetic-based ImageReward as used in Hyper-SD.

In summary, our main contributions are summarized as follows:
\begin{enumerate}[label=\arabic*), wide, labelindent=0pt, leftmargin=*]
    \item We propose Hyper-Bagel, a unified acceleration framework designed to simultaneously speed up both multimodal understanding and generation tasks. Our approach uses a divide-and-conquer strategy, employing speculative decoding for next-token prediction and a multi-stage distillation process for diffusion denoising, ensuring that acceleration in one domain does not compromise the other.

    \item The framework delivers substantial performance gains, achieving over a \textbf{2x} speedup in multimodal understanding. For generative tasks, it yields even more significant acceleration, speeding up text-to-image generation by \textbf{16.67x} and image editing by \textbf{22x}, all while preserving the high-quality output of the original model.

    \item We further develop a highly efficient 1-NFE model that enables near real-time interactive editing and generation. By combining advanced adversarial distillation with human feedback learning, this model achieves ultimate cost-effectiveness and responsiveness, making complex multimodal interactions seamless and instantaneous.
\end{enumerate}

\section{Data}


The VLM image-text paired, text-to-image generation, image-to-image editing and interleaved data used for draft model training and diffusion distillation in this paper are all originated from open-source datasets available online.
\begin{itemize}
    \item \textbf{VLM Image-Text Paired Data}: We utilize the single-image stage data mixture from LLaVA-OneVision~\cite{li2024llava} as our training data for VLM tasks, which includes around 4 million image-text pairs. 
    To align with the prediction distribution of the target model and train a draft model, we also call the target model to generate new answers as in EAGLE3 for each question instead of using the answers from the dataset.
    \item \textbf{Text-to-Image Generation Data}: We incorporate JourneyDB~\cite{sun2023journeydb} as the training data for text-to-image diffusion distillation, which is a synthetic dataset comprising over 4 million images generated by Midjourney along with detailed captions annotated by VLM. 
    Since the original prompts for the images are all provided by users, this synthetic distribution closely aligns with practical usage, and its rich diversity also benefits distillation training.
    \item \textbf{Image-to-Image Editing \& Interleaved Data}: To preserve the emerging properties of BAGEL, we also employ interleaved data into the distillation training process. Specifically, we utilize editing data from Part-2 and Part-3 of the SEED-Data-Edit~\cite{ge2024seed} dataset. Part-2 consists of 52K editing image pairs in real-world scenarios, while Part-3 contains 21K human-annotated multi-turn rounds with up to 5 turns each, totaling 95K editing image pairs.
\end{itemize}
\section{Approach}


\begin{figure}
    \centering
    \includegraphics[width=0.8\linewidth]{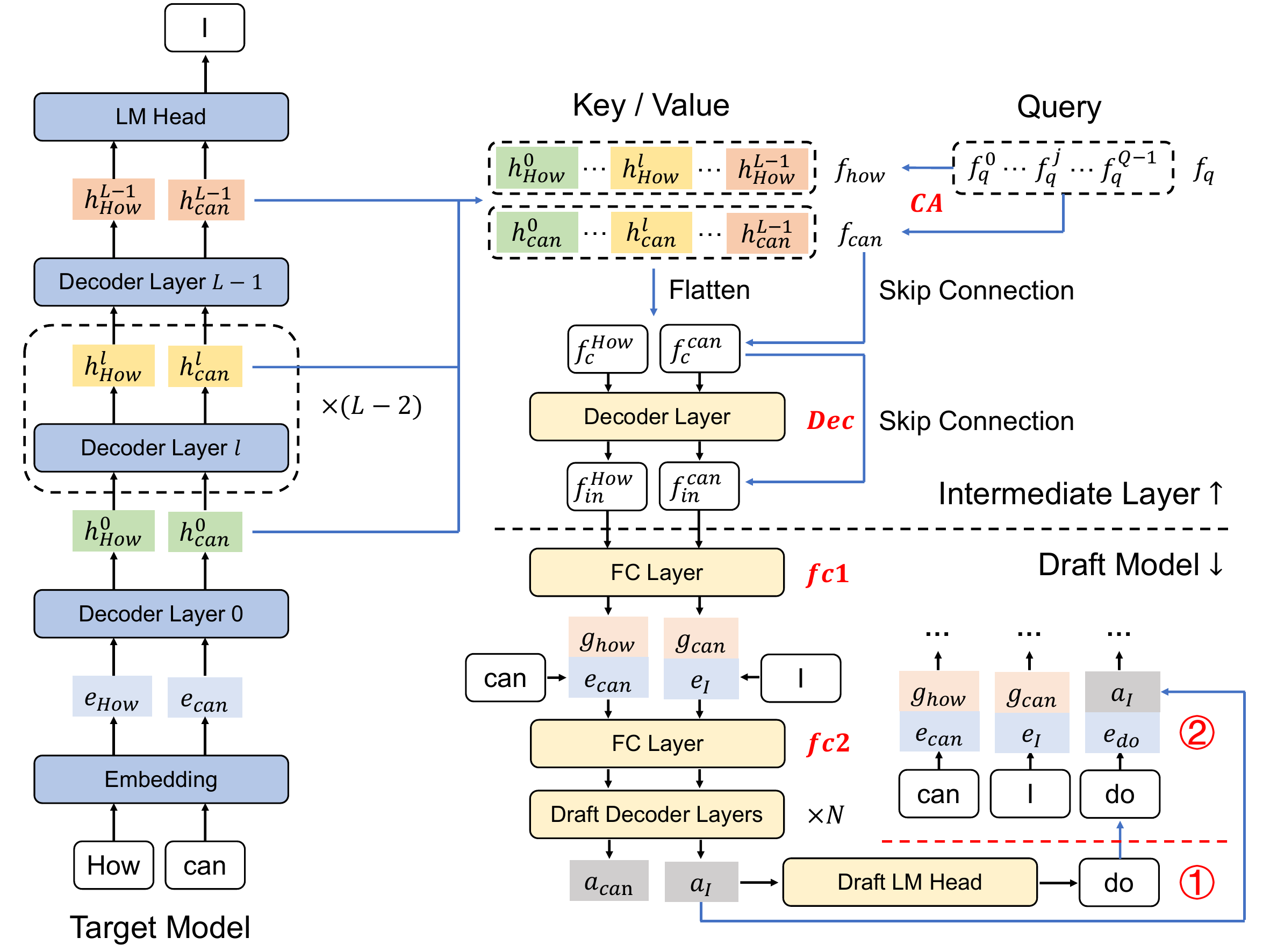}
    \caption{Training pipeline for our proposed speculative decoding approach in Hyper-Bagel.}
    \label{fig:spec}
\end{figure}

\subsection{Speculative Decoding}

In adopting speculative decoding, we primarily adhere to the EAGLE-3~\cite{li2025eagle3scalinginferenceacceleration} training paradigm.
To address the challenge where the differences between different modality tokens make it difficult for draft models with limited capacity to encode target features, we have specifically implemented a series of improvements in the intermediate layer, initialization strategy and loss function to boost EAGLE-3.
The architecture of proposed intermediate layer is illustrated in \cref{fig:spec}, in which we also scale the number of draft decoder layers to $N=2$ to enhance representation capacity.

\subsubsection{Target Feature Aggregation with Meta Queries}
\label{sec:attn}

The intermediate layer of EAGLE~\cite{li2024eagle, li2024eagle2, li2025eagle3scalinginferenceacceleration} plays a crucial role in providing necessary information for the draft model to perform next token prediction, but as we discussed in introduction, it presents greater challenges than others for BAGEL, as much more multimodal tokens including text, ViT tokens, clean latent and noisy latent are all interleaved together.
Our first improvement is to aggregate more features from target model via an attention mechanism rather than simple fully connected layers, which limits the possibility of integrating more feature layers.

Specifically, we initialize several learnable embeddings $f_{q}\in\mathbb{R}^{Q\times D}$ as meta queries, which interact with the features of all target layers for each token $f_i\in\mathbb{R}^{L\times D}$ in the sequence as keys and values through cross-attention, where $i\in[0,S)$, $S$ is the length of input sequence, $L$ is the number of target model layers and $D$ is the feature dimension.
Denote the flatten attention output vectors as $F_c=[f_c^0; f_c^1;\ldots;f_c^{S-1}]\in\mathbb{R}^{S\times QD}$.
Each of them $f_c^i\in\mathbb{R}^{1\times QD}$ is calculated as,
\begin{equation}
    f_c^i=\text{Flatten}(\text{Cross-Attention}(f_q,f_i,f_i)),
\end{equation}
where $f_q$ shares across different tokens in sequence and we use \verb|CA| to represent cross-attention module in the following. 
The sequence $F_c$ then continues forward through a Transformer decoder layer \verb|Dec| with the same architecture as the target model to further aggregate information, finalizing the input $F_{in}\in\mathbb{R}^{S\times D}$ for draft model. 
Note that at this intermediate stage, the input features are not concatenated with token embeddings, and this decoder layer is independent of the ones in the draft model that participate in iterative decoding.

\subsubsection{Zero-Init Fully Connected Layers with Residuals}
\label{sec:zero}

Regarding initialization strategy, our experiments indicate that utilizing the last few layers and the language modeling head of pretrained target model contributes to final performance.
This is based on a fundamental intuition that we aim to reduce the training difficulty of the draft model.
Thus we hope the pre-trained weights can guide the updates of the two most critical fully connected layers through gradient backpropagation, namely \verb|fc1|, which reduces dimensionality in the intermediate layer, and \verb|fc2|, which incorporates token embedding through concatenation at each iterative decoding step.

To achieve this, in addition to leveraging pre-trained weights, we make the following improvements:
\textbf{(1)} zero-init the last projection layer of both \verb|CA| and \verb|Dec| in the intermediate layer;
\textbf{(2)} add skip connections for both \verb|CA| and \verb|Dec|, where we specify low, middle, and high-level feature sequences from the forward pass of target model as residuals similarly to EAGLE-3;
\textbf{(3)} zero-init the \verb|fc1| and \verb|fc2|.

In this way, at the very beginning of training with zero input, the gradients backpropagated from the loss function to \verb|fc2| remain consistent with the target model. 
This means \verb|fc2| needs to aggregate existing information to simulate the feature sequence of the target model at corresponding position. 
When the draft model contains only one decoder layer, \verb|fc2| needs to simulate the output of the second-to-last layer; if the draft has two decoder layers, \verb|fc2| simulates the output of the third-to-last layer.
And the input features initially obtained by \verb|fc1| through skip connections are the same as the three different levels of features in EAGLE-3. 

\subsubsection{Diminish Forward KL Divergence Supervision}
\label{sec:ce}

A prospective issue is that while the soft labels from the target model's probability distribution contain rich knowledge, using forward KL divergence as the loss function to cover all modes may be overly challenging for a draft model with very limited capacity.
To relax this constraint, we additionally introduced a cross-entropy loss supervised by the one-hot hard labels output by the target model,
\begin{equation}
    L_{total}=KL(p_{\text{target}}\|p_{\text{draft}}) + \lambda \cdot [-\log p_{\text{draft}}^{\arg\max(p_{\text{target}})}],
\end{equation}
where $p_{\text{draft}}$ and $p_{\text{target}}$ represent the output distribution probability vectors of draft and target models, respectively, and $p_{*}^k$ refers to the probability of the $k$-th class within the probability distribution $p_*$.
In practice, we set $\lambda=0.1$ and find it yields better performance than vanilla forward KL utilized in EAGLE-3.

\subsection{Diffusion Distillation}

Regarding diffusion distillation, our primary objective is to preserve the model’s full capabilities, including both control and quality perspectives. 
In this paper, we train a 6-NFE model that is lossless across image generation and editing benchmarks, and a highly cost-effective 1-NFE model.
The 6-NFE model is obtained through three-stage training involving CFG Distillation (\cref{sec:stage1}), TSCD (\cref{sec:stage2}) and DMDO (\cref{sec:stage3}).
The 1-NFE model is further fine-tuned from the 6-NFE model through two additional stages, i.e. ADP (\cref{sec:stage4}) and ReFL (\cref{sec:stage5}).

\subsubsection{Stage-1: CFG Distillation}\label{sec:stage1}

To preserve the model's ability to control the degree of instruction adherence and consistency maintenance with the original image in the editing scenario through CFG, we distill CFG embeddings into its single forward pass during the first-stage training.
We curate two additional timestep encoding layers for injecting text scale and image scale, respectively.
While text scale is generic for both image generation and editing, image scale is only used in editing scenarios.
Their architecture and injection positions are completely consistent with the timestep, which is similar to the best practices in FLUX~\cite{flux2024, labs2025flux1kontextflowmatching}, ensuring that control signals can be precisely propagated to every layer of DiT~\cite{peebles2023scalable}.

During the distillation training process, for text-to-image data, we randomly select a text scale value ranging from 1 to 5, while for editing samples, we additionally sample a random image scale value between 1.0 and 2.5 as input.
Regarding the training timestep schedule, we found that setting the diffusion timestep shift to at least 3.0 is necessary to achieve higher structural integrity and image fidelity.

\subsubsection{Stage-2: Trajectory Segmented Consistency Distillation}\label{sec:stage2}

After CFG distillation, we conduct a similar consistency distillation approach as in our previous work Hyper-SD.
Although our goal is a 6-NFE model, we do not adopt progressive distillation like Hyper-SD (e.g., reducing segments from 8$\rightarrow$4$\rightarrow$2$\rightarrow$1), but instead directly implement a 3-segment configuration in one stage.
And we discard the Mean Squared Error (MSE) loss function and shifted entirely to adversarial loss.
In the discriminator design, we adopt the same multi-head architecture as in DMDX to enhance discriminative capability, and the parameters of the pre-trained backbone are also set to be trainable.

The intuition behind all these improvements is to enhance structural integrity in this second stage, while delegating the task of improving image fidelity to the third-stage score distillation.
Eliminating multi-stage progressive distillation simplifies training procedures and reduces training time, while pure adversarial loss and multi-head discriminator design enable the model to focus more on the overall image composition across multiple scales.

\begin{figure}
    \centering
    \includegraphics[width=\linewidth]{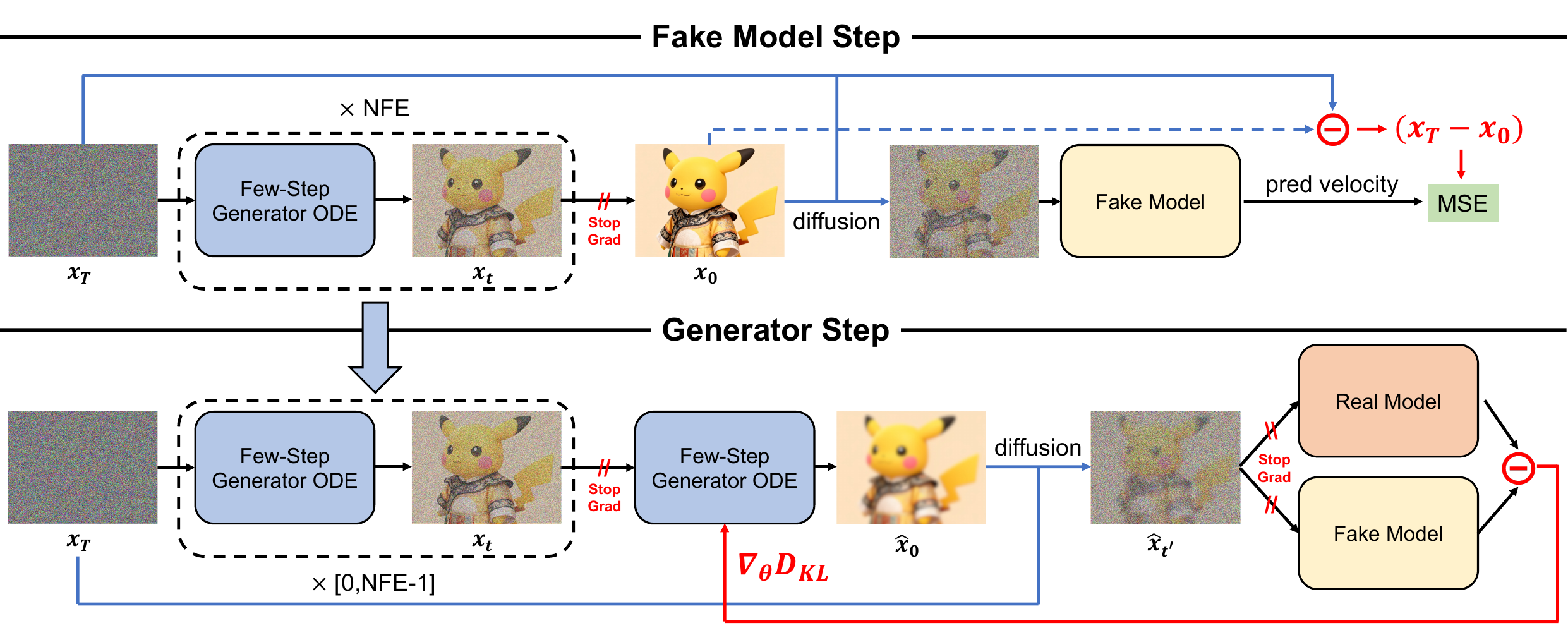}
    \caption{Training pipeline for our proposed Distribution Matching Distillation via ODE (DMDO).}
    \label{fig:dmdo}
\end{figure}

\subsubsection{Stage-3: Distribution Matching Distillation via ODE}\label{sec:stage3}

A major drawback of the DMD series methods is the use of an SDE-based consistency sampler in the few-step generator, which causes the generated images to become overly smooth and lack detail. This contradicts our goal of enhancing image fidelity in the third stage.
For this purpose, we propose Distribution Matching Distillation via ODE (DMDO), aiming to keep the original sampler unchanged and maintain the alignment of the ODE trajectories between the student and teacher models as much as possible.

Specifically, in the fake model update step, we randomly initialize a noise $x_T$ and utilize the ODE of few-step generator to obtain a complete trajectory. 
Taking 6-NFE as an example, we save the complete trajectory $x_T\rightarrow x_{5T/6}\rightarrow x_{4T/6}\rightarrow x_{3T/6}\rightarrow x_{2T/6}\rightarrow x_{1T/6}\rightarrow x_0$ and perform linear interpolation between $x_T$ and $x_0$ to obtain $x_t$ as input for the fake model. 
Therefore, the target of the fake model also transforms into the velocity $(x_T - x_0)$.
This is to enable the fake model to better capture the practical distribution of the few-step generator at each timestep, eliminating the potential distribution shifts introduced by the original practice of adding random noise in DMD.

In the few-step generator update step, we reuse the existing trajectories from the fake update step in order to help reduce training costs and shorten the time for each iteration.
Taking a sampled timestep $t$ falling within the interval between $4T/6$ and $3T/6$ as an example.
We extract the noisy latent $x_{4T/6}$ from the trajectory, and then forward it again through the ODE of the few-step generator with gradients enabled. 
After obtaining a predicted $\hat{x}_0$ at the $4T/6$ timestep, we resample a new timestep $t^\prime$ and perform linear interpolation with $x_T$ to obtain $\hat{x}_{t^\prime}$ serve as an score function input for the fake and real models. 
The subsequent loss function calculation follows the same approach as in DMD, as illustrated in \cref{fig:dmdo}.

Through this alternating optimization of the fake model and the few-step generator, we ultimately obtain a 6-NFE model that is lossless across all benchmark dimensions. 
During this score distillation stage, the model significantly improves the fidelity of generated images, achieving high approximation to the original model in both color vibrancy and detail richness.

\subsubsection{Stage-4: Adversarial Diffusion Pre-training}\label{sec:stage4}

To further achieve 1-NFE image generation and editing for ultimate cost-effectiveness, we additionally introduce the fourth and fifth stages of fine-tuning based on the 6-NFE model. 
We adopt a similar approach to that used in training the 6-NFE model when building the 1-NFE model: first enhancing structural integrity to establish the overall composition of the image, and then refining the colors and details of the generated content. 
However, considering that the model capacity decreases significantly at 1-NFE, it is unrealistic for the original training paradigm to require alignment with the teacher model’s distribution. 
Our approach is to first attempt structural-level alignment with the 6-NFE model’s distribution, and then compensate for deficiencies in fidelity through human feedback.

As for structural integrity in the 1-NFE model, we incorporate an adversarial approach based on rectified flow from DMDX, namely Adversarial Diffusion Pre-training (ADP). 
This approach utilizes the 6-NFE model to sample an ODE trajectory, then performs linear interpolation between the starting point $x_{T}$ and the ending point $x_{0}$ to obtain a noisy latent $x_{t}$, which is subsequently fed into the 1-NFE generator for prediction. 
The $\hat{x}_{0}$ predicted by the 1-NFE generator is evaluated as a fake sample by two different discriminators in the latent space and pixel space, respectively, with the real target sample being the ODE trajectory endpoint $x_{0}$ itself. This method aligns perfectly with our training objectives because we only aim to achieve alignment with the 6-NFE model, and the cost of sampling ODE trajectories using the 6-NFE model is relatively low.

\subsubsection{Stage-5: Reward Feedback Learning}\label{sec:stage5}

Regarding image fidelity for the 1-NFE model, we follow the common practice in Hyper-SD to introduce human feedback.
Unlike previous approaches~\cite{zhang2024unifl}, we do not adopt the training paradigm of multiple reward models.
Instead, we use only a more comprehensive VLM-based reward model to provide guidance.
This is considering that BAGEL is also a multimodal model, with its capabilities focused on semantic understanding of prompts. 
Therefore, a reward model based on VLM is highly suitable, as VLM can scale to a larger capacity, enabling richer knowledge, and the scaling of reward models has been proven to be highly effective in RewardDance~\cite{wu2025rewarddance}.
The loss function we use in ReFL is consistent with the aesthetic supervision loss function in Hyper-SD, where the reward activation threshold $\alpha_d$ for the ReLU function is set to 6.0.

\section{Experiments}


\begin{table}[t]
    \centering
    \small
    \begin{tabular}{lccc}
        \toprule[2pt]
        \textbf{Method} & \textbf{Average Acceptance Length} & \textbf{Acceptance Rate 10-$\bm\alpha$} \\
        \midrule[1pt]
        Vanilla EAGLE-3 & 3.6184 & 0.7327 \\
        \rowcolor{seedblue!10} \textbf{Hyper-Bagel (Ours)} & 3.7709 & 0.7452 \\
        - w/o Zero-Init & 2.8273 & 0.6494 \\
        - w/o CE Loss & 3.6642 & 0.7365 \\
        - w/o Zero-Init \& CE Loss & 3.4832 & 0.7207 \\
        \bottomrule[2pt]
    \end{tabular}
    \caption{Quantitative results in speculative decoding.}
    \label{tab:spec}
\end{table}

\subsection{Setup}

During the diffusion distillation process, we freeze the parameters of the understanding branch.
Therefore, since the draft model trained through speculative decoding undergoes verification by the target model after prediction, the performance metrics of understanding remain unaffected. 
We follow EAGLE-3~\cite{li2025eagle3scalinginferenceacceleration} in reporting the average acceptance length of the draft model.
For generation benchmarks, we follow BAGEL~\cite{deng2025emerging} to report the performance of GenEval and GEdit-Bench for image generation and editing tasks respectively.

Some might worry whether diffusion distillation affects text generation, especially the thinking capability in interleaved scenarios, since VAE tokens are incorporated into the sequence context.
However, in the context management of BAGEL, the clean latent after each diffusion denoising is incorporated into the context via prefill, without retaining the KV cache of the noisy latent. 
This means that during deployment, we can separately deploy the generation branch weights of the distilled model and the original model.
We only invoke the distilled model during denoising and revert to the original model during prefilling, ensuring no loss in understanding performance.

\subsection{Quantitative Results}

\begin{table}[t]
    \centering
    \small
    \begin{tabular}{lccccccc}
        \toprule[2pt]
        \textbf{Model} & \textbf{Single Obj.} & \textbf{Two Obj.} & \textbf{Counting} & \textbf{Colors} & \textbf{Position} & \textbf{Color Attri.} & \textbf{Overall$\uparrow$} \\
        \midrule[1pt]
        Chameleon$^*$~\cite{team2024chameleon} & - & - & - & - & - & - & 0.39 \\
        LWM$^*$~\cite{liu2024world} & 0.93 & 0.41 & 0.46 & 0.79 & 0.09 & 0.15 & 0.47 \\
        SEED-X$^*$~\cite{ge2024seed} & 0.97 & 0.58 & 0.26 & 0.80 & 0.19 & 0.14 & 0.49 \\
        TokenFlow-XL$^*$~\cite{qu2024tokenflow} & 0.95 & 0.60 & 0.41 & 0.81 & 0.16 & 0.24 & 0.55 \\
        ILLUME$^*$~\cite{wang2024illume} & 0.99 & 0.86 & 0.45 & 0.71 & 0.39 & 0.28 & 0.61 \\
        Janus$^*$~\cite{wu2025janus} & 0.97 & 0.68 & 0.30 & 0.84 & 0.46 & 0.42 & 0.61 \\
        Transfusion$^*$~\cite{zhou2024transfusion} & - & - & - & - & - & - & 0.63 \\
        Emu3-Gen$^*$~\cite{sun2023emu} & 0.99 & 0.81 & 0.42 & 0.80 & 0.49 & 0.45 & 0.66 \\
        Show-o$^*$~\cite{xie2024show} & 0.98 & 0.80 & 0.66 & 0.84 & 0.31 & 0.50 & 0.68 \\
        Janus-Pro-7B$^*$~\cite{chen2025janus} & 0.99 & 0.89 & 0.59 & 0.90 & 0.79 & 0.66 & 0.80 \\
        MetaQuery-XL$^*$~\cite{pan2025transfer} & - & - & - & - & - & - & 0.80 \\
        \midrule[1pt]
        \textcolor{gray}{BAGEL$^*$ (100-NFE)}~\cite{deng2025emerging} & \textcolor{gray}{0.98} & \textcolor{gray}{0.95} & \textcolor{gray}{0.84} & \textcolor{gray}{0.95} & \textcolor{gray}{0.78} & \textcolor{gray}{0.77} & \textcolor{gray}{0.88} \\
        BAGEL$^\dagger$ (100-NFE)~\cite{deng2025emerging} & 0.9875 & 0.9520 & 0.8281 & 0.9415 & 0.7400 & 0.7350 & \underline{0.8640} \\
        \rowcolor{seedblue!10}\textbf{Hyper-BAGEL (6-NFE)} & 0.9938 & 0.9394 & 0.8562 & 0.9388 & 0.7175 & 0.7425 & \textbf{0.8647} \\
        \rowcolor{seedblue!10}\textbf{Hyper-BAGEL (1-NFE)} & 0.9719 & 0.8586 & 0.7500 & 0.9043 & 0.6725 & 0.6200 & 0.7962 \\
        \bottomrule[2pt]
    \end{tabular}
    \caption{Quantitative results on GenEval. $^*$ Results are cited from those reported in the BAGEL paper. $^\dagger$ We reproduce these results under the same training environment.}
    \label{tab:geneval}
\end{table}

\begin{table}[t]
    \centering
    \small
    \begin{tabular}{lcccccc}
        \toprule[2pt]
        \multirow{2}[1]{*}{\textbf{Model}} & \multicolumn{3}{c}{\textbf{GEdit-Bench-EN (Full set)}} & \multicolumn{3}{c}{\textbf{GEdit-Bench-CN (Full set)}} \\
        \cmidrule(lr){2-4} \cmidrule(lr){5-7}
         & \textbf{G\_SC} & \textbf{G\_PQ} & \textbf{G\_O} & \textbf{G\_SC} & \textbf{G\_PQ} & \textbf{G\_O} \\
        \midrule[1pt]

        Instruct-Pix2Pix$^*$~\cite{brooks2023instructpix2pix} & 3.58 & 5.49 & 3.68 & - & - & - \\
        MagicBrush$^*$~\cite{zhang2023magicbrush} & 4.68 & 5.66 & 4.52 & - & - & - \\
        AnyEdit$^*$~\cite{yu2025anyedit} & 3.18 & 5.82 & 3.21 & - & - & - \\
        OmniGen$^*$~\cite{xiao2025omnigen} & 5.96 & 5.89 & 5.06 & - & - & - \\
        Step1X-Edit$^*$~\cite{liu2025step1x} & 7.09 & 6.76 & \textbf{6.70} & 7.20 & 6.87 & \textbf{6.86} \\

        \midrule[1pt]
        
        \textcolor{gray}{BAGEL$^*$ (132-NFE)}~\cite{deng2025emerging} & \textcolor{gray}{7.36} & \textcolor{gray}{6.83} & \textcolor{gray}{6.52} & \textcolor{gray}{7.34} & \textcolor{gray}{6.85} & \textcolor{gray}{6.50} \\
        BAGEL$^\dagger$ (132-NFE)~\cite{deng2025emerging} & 7.506 & 6.808 & 6.602 & 7.533 & 6.849 & 6.610 \\
        \rowcolor{seedblue!10}\textbf{Hyper-BAGEL (6-NFE)} & 7.448 & 6.757 & \underline{6.612} & 7.568 & 6.808 & \underline{6.671} \\
        \rowcolor{seedblue!10}\textbf{Hyper-BAGEL (1-NFE)} & 7.069 & 5.969 & 5.975 & 6.997 & 6.000 & 5.966 \\

        \bottomrule[2pt]
    \end{tabular}
    \caption{Quantitative results on GEdit-Bench. The baseline is 132-NFE because it uses a CFG interval of [0.4, 1.0], where text and image CFG are simultaneously enabled only at timesteps within this interval. $^*$ Results are cited from those reported in the BAGEL paper. $^\dagger$ We reproduce these results under the same training environment.}
    \label{tab:gedit}
\end{table}

\textbf{Speculative decoding.}
In \cref{tab:spec}, we present the average acceptance length $\tau$ and acceptance rate 10-$\alpha$ with 10 extrapolation steps using chain decoding for multiple EAGLE-3 variants including ours.
Our full Hyper-Bagel framework achieves the best performance, outperforming the Vanilla EAGLE-3 baseline. 
The results show that removing the zero-initialization strategy causes the most significant performance degradation, highlighting its critical role in bridging the target and draft models. 
While the cross-entropy loss also proves beneficial, its absence leads to a smaller decline. 
Interestingly, the model performs better without both components than it does when only zero-initialization is removed, suggesting that the strict cross-entropy loss constraint may become counterproductive without the foundational alignment provided by zero-initialization, thus hindering the training process.

\textbf{Image generation.}
As shown in Table~\ref{tab:geneval}, our 6-NFE Hyper-BAGEL model demonstrates lossless performance on the GenEval benchmark. It achieves an overall score of 0.8647, slightly surpassing the 0.8640 score of the 100-NFE BAGEL baseline. This result confirms that our distillation process achieves a significant 16.67x speedup by reducing sampling steps from 100 to 6 NFE without compromising generative quality. Furthermore, our highly efficient 1-NFE model remains competitive with other state-of-the-art unified models, as its overall score of 0.7962 is on par with leading models like Janus-Pro-7B and MetaQuery-XL, which both score 0.80.

\textbf{Image editing.}
The lossless nature of our accelerated model is evident in the image editing tasks evaluated on GEdit-Bench (Table~\ref{tab:gedit}). 
Our 6-NFE Hyper-BAGEL consistently outperforms the 132-NFE baseline across both English and Chinese datasets, posting higher overall scores of 6.612 and 6.671 respectively. 
This remarkable consistency is maintained despite a massive reduction in computational cost, translating to an approximately 22x inference speedup. 
Notably, our 1-NFE model, designed for maximum efficiency, still delivers strong performance, with its overall scores of 5.975 in English and 5.966 in Chinese significantly surpassing established methods like OmniGen.

\subsection{Qualitative Results}

\begin{figure}
    \centering
    \includegraphics[width=\linewidth]{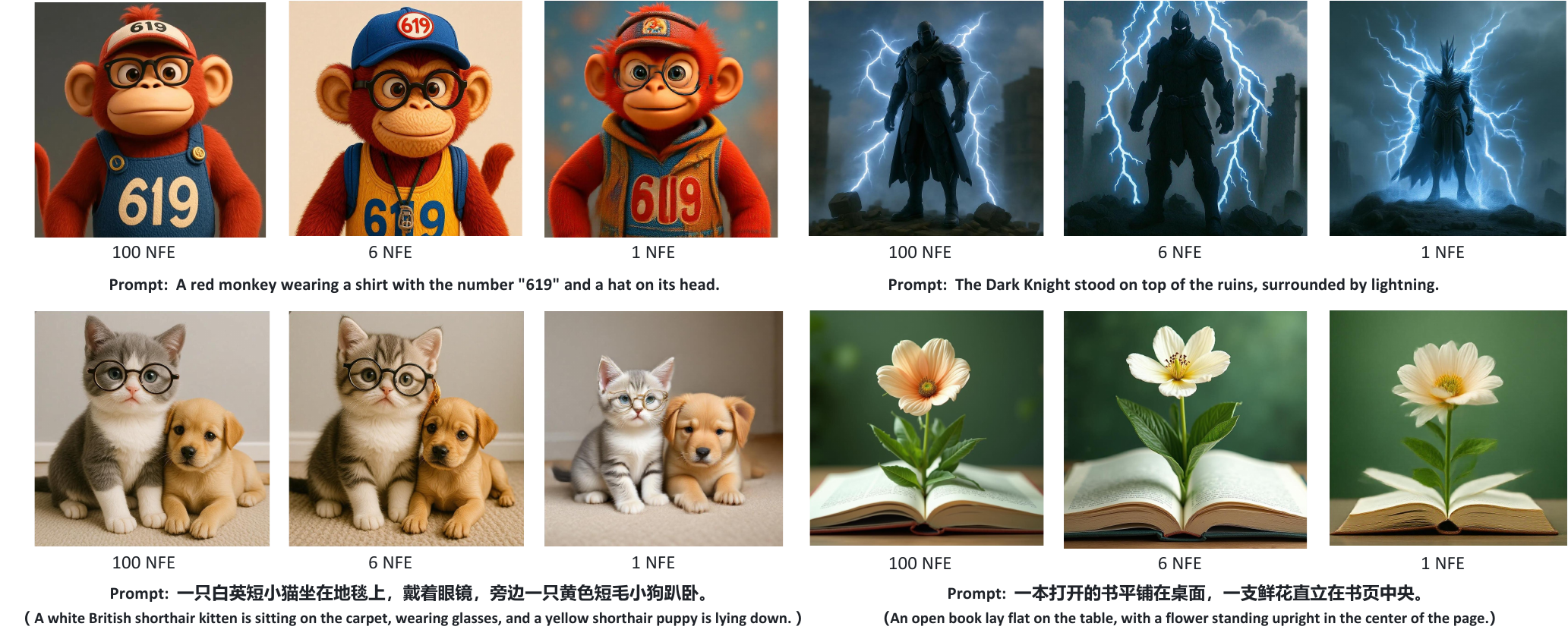}
    \caption{Qualitative comparison of different accelerated models against the baseline on image generation.}
    \label{fig:qualitative_t2i}
\end{figure}

\begin{figure}
    \centering
    \includegraphics[width=\linewidth]{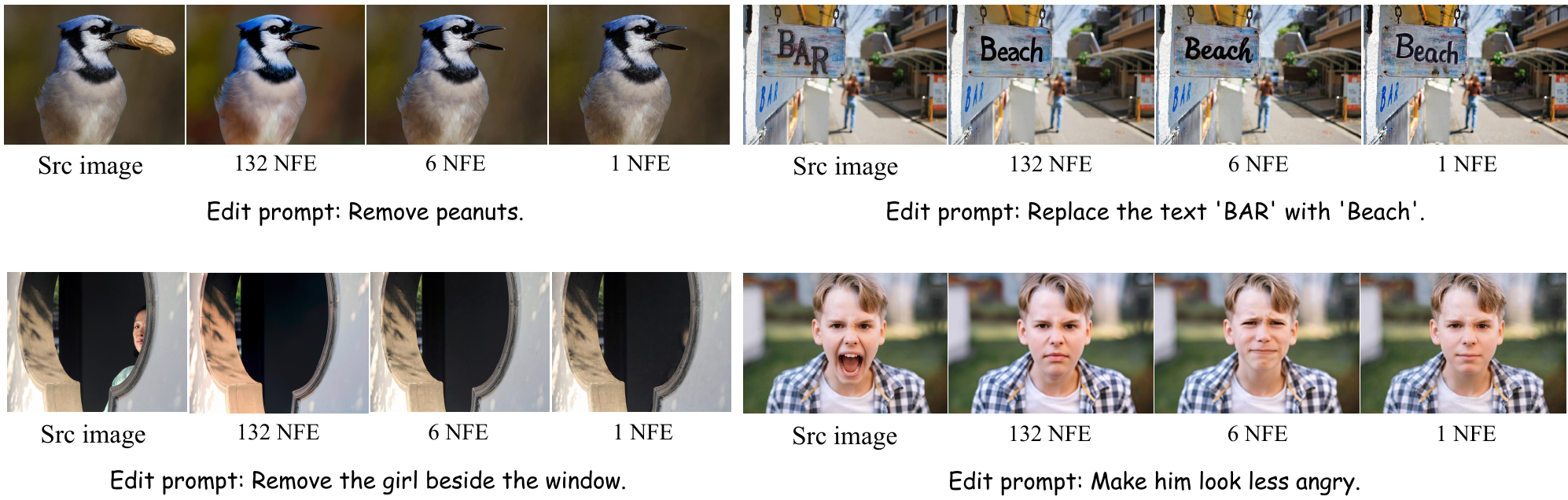}
    \caption{Qualitative comparison of different accelerated models against the baseline on image editing.}
    \label{fig:qualitative_edit}
\end{figure}

\textbf{Image generation.}
The \cref{fig:qualitative_t2i} presents a qualitative comparison of our accelerated models against the 100-NFE baseline on the image generation task. 
It is clear from all four cases that the results from our 6-NFE model are visually indistinguishable from the baseline, faithfully reproducing intricate details like the number "619" on the monkey's shirt, the complex composition and lighting in the Dark Knight scene, and the texture of the animals' fur. 
In contrast, the 1-NFE model, designed for maximum efficiency, demonstrates a clear trade-off.
While it rapidly generates images that are highly relevant to the prompt, it does so with a reduction in detail fidelity. 
For instance, it may occasionally omit key elements from the prompt, such as the glasses on the kitten, or exhibit minor deviations in detail, like the number on the monkey's shirt. 
Nevertheless, the core semantics and overall quality of its outputs remain highly competitive. 
These visual comparisons strongly validate that our 6-NFE model achieves lossless acceleration, while the 1-NFE model serves as an efficient and reliable option for applications where real-time interaction is prioritized.

\textbf{Image editing.}
The \cref{fig:qualitative_edit} illustrates the qualitative performance of our accelerated models compared to the 132-NFE baseline in image editing tasks. 
For each case, the 6-NFE Hyper-BAGEL model demonstrates exceptional fidelity, executing precise edits, such as removing peanuts, replacing text, or eliminating a person, with results virtually indistinguishable from the high-NFE baseline. 
This confirms the lossless nature of our 6-NFE acceleration, providing a significant speedup without any perceptible degradation in editing quality. 
A notable advantage emerges with the 1-NFE model in the editing context.
Its ability to leverage the structural and contextual information from the source image allows it to maintain strong visual coherence and successfully apply the requested edits. 
While subtle details or perfect photographic realism might be slightly compromised compared to the higher-NFE versions, the 1-NFE model still offers highly usable and contextually accurate edits, making it a powerful tool for near real-time interactive editing where the immediate visual feedback of a fast model is invaluable.

\section{Conclusion}
In this work, we introduced Hyper-Bagel, a unified framework designed to successfully mitigate the significant computational overhead in advanced multimodal models. 
Our divide-and-conquer strategy, employing speculative decoding for understanding and multi-stage distillation for generation, has proven highly effective through comprehensive experiments. 
We have demonstrated that our lossless 6-NFE model accelerates text-to-image generation and editing by over 16.67x and 22x respectively, achieving performance on par with or even superior to the high-NFE baseline, while simultaneously doubling the speed of multimodal understanding. 
Furthermore, our highly efficient 1-NFE model stands as a robust and practical solution for near real-time applications, proving especially effective in interactive editing. 
Ultimately, Hyper-Bagel provides a holistic solution that closes the gap between the powerful capabilities of unified multimodal models and the practical demands of real-world deployment, enabling seamless and instantaneous creative interaction without compromise.

\clearpage

\bibliographystyle{plainnat}
\bibliography{main}

\begin{thebibliography}{55}
\providecommand{\natexlab}[1]{#1}
\providecommand{\url}[1]{\texttt{#1}}
\expandafter\ifx\csname urlstyle\endcsname\relax
  \providecommand{\doi}[1]{doi: #1}\else
  \providecommand{\doi}{doi: \begingroup \urlstyle{rm}\Url}\fi

\bibitem[Achiam et~al.(2023)Achiam, Adler, Agarwal, Ahmad, Akkaya, Aleman, Almeida, Altenschmidt, Altman, Anadkat, et~al.]{achiam2023gpt}
Josh Achiam, Steven Adler, Sandhini Agarwal, Lama Ahmad, Ilge Akkaya, Florencia~Leoni Aleman, Diogo Almeida, Janko Altenschmidt, Sam Altman, Shyamal Anadkat, et~al.
\newblock Gpt-4 technical report.
\newblock \emph{arXiv preprint arXiv:2303.08774}, 2023.

\bibitem[Brooks et~al.(2023)Brooks, Holynski, and Efros]{brooks2023instructpix2pix}
Tim Brooks, Aleksander Holynski, and Alexei~A Efros.
\newblock Instructpix2pix: Learning to follow image editing instructions.
\newblock In \emph{Proceedings of the IEEE/CVF conference on computer vision and pattern recognition}, pages 18392--18402, 2023.

\bibitem[Brown et~al.(2020)Brown, Mann, Ryder, Subbiah, Kaplan, Dhariwal, Neelakantan, Shyam, Sastry, Askell, et~al.]{brown2020language}
Tom Brown, Benjamin Mann, Nick Ryder, Melanie Subbiah, Jared~D Kaplan, Prafulla Dhariwal, Arvind Neelakantan, Pranav Shyam, Girish Sastry, Amanda Askell, et~al.
\newblock Language models are few-shot learners.
\newblock \emph{Advances in neural information processing systems}, 33:\penalty0 1877--1901, 2020.

\bibitem[Cai et~al.(2024)Cai, Li, Geng, Peng, Lee, Chen, and Dao]{cai2024medusa}
Tianle Cai, Yuhong Li, Zhengyang Geng, Hongwu Peng, Jason~D Lee, Deming Chen, and Tri Dao.
\newblock Medusa: Simple llm inference acceleration framework with multiple decoding heads.
\newblock \emph{arXiv preprint arXiv:2401.10774}, 2024.

\bibitem[Chen et~al.(2025)Chen, Wu, Liu, Pan, Liu, Xie, Yu, and Ruan]{chen2025janus}
Xiaokang Chen, Zhiyu Wu, Xingchao Liu, Zizheng Pan, Wen Liu, Zhenda Xie, Xingkai Yu, and Chong Ruan.
\newblock Janus-pro: Unified multimodal understanding and generation with data and model scaling.
\newblock \emph{arXiv preprint arXiv:2501.17811}, 2025.

\bibitem[Deng et~al.(2025)Deng, Zhu, Li, Gou, Li, Wang, Zhong, Yu, Nie, Song, et~al.]{deng2025emerging}
Chaorui Deng, Deyao Zhu, Kunchang Li, Chenhui Gou, Feng Li, Zeyu Wang, Shu Zhong, Weihao Yu, Xiaonan Nie, Ziang Song, et~al.
\newblock Emerging properties in unified multimodal pretraining.
\newblock \emph{arXiv preprint arXiv:2505.14683}, 2025.

\bibitem[Gao et~al.(2025)Gao, Gong, Guo, Hou, Lai, Li, Li, Lian, Liao, Liu, et~al.]{gao2025seedream}
Yu~Gao, Lixue Gong, Qiushan Guo, Xiaoxia Hou, Zhichao Lai, Fanshi Li, Liang Li, Xiaochen Lian, Chao Liao, Liyang Liu, et~al.
\newblock Seedream 3.0 technical report.
\newblock \emph{arXiv preprint arXiv:2504.11346}, 2025.

\bibitem[Ge et~al.(2024)Ge, Zhao, Zhu, Ge, Yi, Song, Li, Ding, and Shan]{ge2024seed}
Yuying Ge, Sijie Zhao, Jinguo Zhu, Yixiao Ge, Kun Yi, Lin Song, Chen Li, Xiaohan Ding, and Ying Shan.
\newblock Seed-x: Multimodal models with unified multi-granularity comprehension and generation.
\newblock \emph{arXiv preprint arXiv:2404.14396}, 2024.

\bibitem[Ghosh et~al.(2023)Ghosh, Hajishirzi, and Schmidt]{ghosh2023geneval}
Dhruba Ghosh, Hannaneh Hajishirzi, and Ludwig Schmidt.
\newblock Geneval: An object-focused framework for evaluating text-to-image alignment.
\newblock \emph{Advances in Neural Information Processing Systems}, 36:\penalty0 52132--52152, 2023.

\bibitem[Grattafiori et~al.(2024)Grattafiori, Dubey, Jauhri, Pandey, Kadian, Al-Dahle, Letman, Mathur, Schelten, Vaughan, et~al.]{grattafiori2024llama}
Aaron Grattafiori, Abhimanyu Dubey, Abhinav Jauhri, Abhinav Pandey, Abhishek Kadian, Ahmad Al-Dahle, Aiesha Letman, Akhil Mathur, Alan Schelten, Alex Vaughan, et~al.
\newblock The llama 3 herd of models.
\newblock \emph{arXiv preprint arXiv:2407.21783}, 2024.

\bibitem[Ho and Salimans(2022)]{ho2022classifier}
Jonathan Ho and Tim Salimans.
\newblock Classifier-free diffusion guidance.
\newblock \emph{arXiv preprint arXiv:2207.12598}, 2022.

\bibitem[Ho et~al.(2020)Ho, Jain, and Abbeel]{ho2020denoising}
Jonathan Ho, Ajay Jain, and Pieter Abbeel.
\newblock Denoising diffusion probabilistic models.
\newblock \emph{Advances in neural information processing systems}, 33:\penalty0 6840--6851, 2020.

\bibitem[Labs(2024)]{flux2024}
Black~Forest Labs.
\newblock Flux.
\newblock \url{https://github.com/black-forest-labs/flux}, 2024.

\bibitem[Labs et~al.(2025)Labs, Batifol, Blattmann, Boesel, Consul, Diagne, Dockhorn, English, English, Esser, Kulal, Lacey, Levi, Li, Lorenz, Müller, Podell, Rombach, Saini, Sauer, and Smith]{labs2025flux1kontextflowmatching}
Black~Forest Labs, Stephen Batifol, Andreas Blattmann, Frederic Boesel, Saksham Consul, Cyril Diagne, Tim Dockhorn, Jack English, Zion English, Patrick Esser, Sumith Kulal, Kyle Lacey, Yam Levi, Cheng Li, Dominik Lorenz, Jonas Müller, Dustin Podell, Robin Rombach, Harry Saini, Axel Sauer, and Luke Smith.
\newblock Flux.1 kontext: Flow matching for in-context image generation and editing in latent space, 2025.
\newblock URL \url{https://arxiv.org/abs/2506.15742}.

\bibitem[Leviathan et~al.(2023)Leviathan, Kalman, and Matias]{leviathan2023fast}
Yaniv Leviathan, Matan Kalman, and Yossi Matias.
\newblock Fast inference from transformers via speculative decoding.
\newblock In \emph{International Conference on Machine Learning}, pages 19274--19286. PMLR, 2023.

\bibitem[Li et~al.(2024{\natexlab{a}})Li, Zhang, Guo, Zhang, Li, Zhang, Zhang, Zhang, Li, Liu, et~al.]{li2024llava}
Bo~Li, Yuanhan Zhang, Dong Guo, Renrui Zhang, Feng Li, Hao Zhang, Kaichen Zhang, Peiyuan Zhang, Yanwei Li, Ziwei Liu, et~al.
\newblock Llava-onevision: Easy visual task transfer.
\newblock \emph{arXiv preprint arXiv:2408.03326}, 2024{\natexlab{a}}.

\bibitem[Li et~al.(2022)Li, Xia, Li, Li, Wang, Xiao, Wang, Zheng, and Pan]{li2022next}
Jiashi Li, Xin Xia, Wei Li, Huixia Li, Xing Wang, Xuefeng Xiao, Rui Wang, Min Zheng, and Xin Pan.
\newblock Next-vit: Next generation vision transformer for efficient deployment in realistic industrial scenarios.
\newblock \emph{arXiv preprint arXiv:2207.05501}, 2022.

\bibitem[Li et~al.(2024{\natexlab{b}})Li, Wei, Zhang, and Zhang]{li2024eagle}
Yuhui Li, Fangyun Wei, Chao Zhang, and Hongyang Zhang.
\newblock {EAGLE}: Speculative sampling requires rethinking feature uncertainty.
\newblock In \emph{International Conference on Machine Learning}, 2024{\natexlab{b}}.

\bibitem[Li et~al.(2024{\natexlab{c}})Li, Wei, Zhang, and Zhang]{li2024eagle2}
Yuhui Li, Fangyun Wei, Chao Zhang, and Hongyang Zhang.
\newblock {EAGLE-2}: Faster inference of language models with dynamic draft trees.
\newblock In \emph{Empirical Methods in Natural Language Processing}, 2024{\natexlab{c}}.

\bibitem[Li et~al.(2025)Li, Wei, Zhang, and Zhang]{li2025eagle3scalinginferenceacceleration}
Yuhui Li, Fangyun Wei, Chao Zhang, and Hongyang Zhang.
\newblock {EAGLE-3}: Scaling up inference acceleration of large language models via training-time test, 2025.
\newblock URL \url{https://arxiv.org/abs/2503.01840}.

\bibitem[Liao et~al.(2025)Liao, Liu, Wang, Luo, Zhang, Zhao, Wu, Li, Tian, and Huang]{liao2025mogao}
Chao Liao, Liyang Liu, Xun Wang, Zhengxiong Luo, Xinyu Zhang, Wenliang Zhao, Jie Wu, Liang Li, Zhi Tian, and Weilin Huang.
\newblock Mogao: An omni foundation model for interleaved multi-modal generation.
\newblock \emph{arXiv preprint arXiv:2505.05472}, 2025.

\bibitem[Lin et~al.(2024)Lin, Wang, and Yang]{lin2024sdxl}
Shanchuan Lin, Anran Wang, and Xiao Yang.
\newblock Sdxl-lightning: Progressive adversarial diffusion distillation.
\newblock \emph{arXiv preprint arXiv:2402.13929}, 2024.

\bibitem[Lin et~al.(2025)Lin, Xia, Ren, Yang, Xiao, and Jiang]{lin2025diffusion}
Shanchuan Lin, Xin Xia, Yuxi Ren, Ceyuan Yang, Xuefeng Xiao, and Lu~Jiang.
\newblock Diffusion adversarial post-training for one-step video generation.
\newblock \emph{arXiv preprint arXiv:2501.08316}, 2025.

\bibitem[Lipman et~al.(2022)Lipman, Chen, Ben-Hamu, Nickel, and Le]{lipman2022flow}
Yaron Lipman, Ricky~TQ Chen, Heli Ben-Hamu, Maximilian Nickel, and Matt Le.
\newblock Flow matching for generative modeling.
\newblock \emph{arXiv preprint arXiv:2210.02747}, 2022.

\bibitem[Liu et~al.(2024{\natexlab{a}})Liu, Feng, Xue, Wang, Wu, Lu, Zhao, Deng, Zhang, Ruan, et~al.]{liu2024deepseek}
Aixin Liu, Bei Feng, Bing Xue, Bingxuan Wang, Bochao Wu, Chengda Lu, Chenggang Zhao, Chengqi Deng, Chenyu Zhang, Chong Ruan, et~al.
\newblock Deepseek-v3 technical report.
\newblock \emph{arXiv preprint arXiv:2412.19437}, 2024{\natexlab{a}}.

\bibitem[Liu et~al.(2024{\natexlab{b}})Liu, Yan, Zaharia, and Abbeel]{liu2024world}
Hao Liu, Wilson Yan, Matei Zaharia, and Pieter Abbeel.
\newblock World model on million-length video and language with blockwise ringattention.
\newblock \emph{arXiv preprint arXiv:2402.08268}, 2024{\natexlab{b}}.

\bibitem[Liu et~al.(2025)Liu, Han, Xing, Yin, Wang, Cheng, Liao, Wang, Fu, Han, et~al.]{liu2025step1x}
Shiyu Liu, Yucheng Han, Peng Xing, Fukun Yin, Rui Wang, Wei Cheng, Jiaqi Liao, Yingming Wang, Honghao Fu, Chunrui Han, et~al.
\newblock Step1x-edit: A practical framework for general image editing.
\newblock \emph{arXiv preprint arXiv:2504.17761}, 2025.

\bibitem[Lu et~al.(2025)Lu, Ren, Xia, Lin, Wang, Xiao, Ma, Xie, and Lai]{lu2025adversarial}
Yanzuo Lu, Yuxi Ren, Xin Xia, Shanchuan Lin, Xing Wang, Xuefeng Xiao, Andy~J Ma, Xiaohua Xie, and Jian-Huang Lai.
\newblock Adversarial distribution matching for diffusion distillation towards efficient image and video synthesis.
\newblock \emph{arXiv preprint arXiv:2507.18569}, 2025.

\bibitem[Ma et~al.(2025)Ma, Wu, Sun, and Li]{ma2025hpsv3widespectrumhumanpreference}
Yuhang Ma, Xiaoshi Wu, Keqiang Sun, and Hongsheng Li.
\newblock Hpsv3: Towards wide-spectrum human preference score, 2025.
\newblock URL \url{https://arxiv.org/abs/2508.03789}.

\bibitem[Pan et~al.(2025)Pan, Shukla, Singh, Zhao, Mishra, Wang, Xu, Chen, Li, Juefei-Xu, et~al.]{pan2025transfer}
Xichen Pan, Satya~Narayan Shukla, Aashu Singh, Zhuokai Zhao, Shlok~Kumar Mishra, Jialiang Wang, Zhiyang Xu, Jiuhai Chen, Kunpeng Li, Felix Juefei-Xu, et~al.
\newblock Transfer between modalities with metaqueries.
\newblock \emph{arXiv preprint arXiv:2504.06256}, 2025.

\bibitem[Peebles and Xie(2023)]{peebles2023scalable}
William Peebles and Saining Xie.
\newblock Scalable diffusion models with transformers.
\newblock In \emph{Proceedings of the IEEE/CVF international conference on computer vision}, pages 4195--4205, 2023.

\bibitem[Qu et~al.(2024)Qu, Zhang, Liu, Wang, Jiang, Gao, Ye, Du, Yuan, and Wu]{qu2024tokenflow}
Liao Qu, Huichao Zhang, Yiheng Liu, Xu~Wang, Yi~Jiang, Yiming Gao, Hu~Ye, Daniel~K Du, Zehuan Yuan, and Xinglong Wu.
\newblock Tokenflow: Unified image tokenizer for multimodal understanding and generation.
\newblock \emph{arXiv preprint arXiv:2412.03069}, 2024.

\bibitem[Ren et~al.(2024)Ren, Wu, Lu, Kuang, Xia, Wang, Wang, Zhu, Xie, Wang, et~al.]{ren2024byteedit}
Yuxi Ren, Jie Wu, Yanzuo Lu, Huafeng Kuang, Xin Xia, Xionghui Wang, Qianqian Wang, Yixing Zhu, Pan Xie, Shiyin Wang, et~al.
\newblock Byteedit: Boost, comply and accelerate generative image editing.
\newblock In \emph{European Conference on Computer Vision}, pages 184--200. Springer, 2024.

\bibitem[Ren et~al.(2025)Ren, Xia, Lu, Zhang, Wu, Xie, Wang, and Xiao]{ren2025hyper}
Yuxi Ren, Xin Xia, Yanzuo Lu, Jiacheng Zhang, Jie Wu, Pan Xie, Xing Wang, and Xuefeng Xiao.
\newblock Hyper-sd: Trajectory segmented consistency model for efficient image synthesis.
\newblock \emph{Advances in Neural Information Processing Systems}, 37:\penalty0 117340--117362, 2025.

\bibitem[Seawead et~al.(2025)Seawead, Yang, Lin, Zhao, Lin, Ma, Guo, Chen, Qi, Wang, et~al.]{seawead2025seaweed}
Team Seawead, Ceyuan Yang, Zhijie Lin, Yang Zhao, Shanchuan Lin, Zhibei Ma, Haoyuan Guo, Hao Chen, Lu~Qi, Sen Wang, et~al.
\newblock Seaweed-7b: Cost-effective training of video generation foundation model.
\newblock \emph{arXiv preprint arXiv:2504.08685}, 2025.

\bibitem[Shao et~al.(2025)Shao, Xia, Yang, Ren, Wang, and Xiao]{shao2025rayflow}
Huiyang Shao, Xin Xia, Yuhong Yang, Yuxi Ren, Xing Wang, and Xuefeng Xiao.
\newblock Rayflow: Instance-aware diffusion acceleration via adaptive flow trajectories.
\newblock \emph{arXiv preprint arXiv:2503.07699}, 2025.

\bibitem[Sun et~al.(2023{\natexlab{a}})Sun, Pan, Ge, Li, Duan, Wu, Zhang, Zhou, Qin, Wang, et~al.]{sun2023journeydb}
Keqiang Sun, Junting Pan, Yuying Ge, Hao Li, Haodong Duan, Xiaoshi Wu, Renrui Zhang, Aojun Zhou, Zipeng Qin, Yi~Wang, et~al.
\newblock Journeydb: A benchmark for generative image understanding.
\newblock \emph{Advances in neural information processing systems}, 36:\penalty0 49659--49678, 2023{\natexlab{a}}.

\bibitem[Sun et~al.(2023{\natexlab{b}})Sun, Yu, Cui, Zhang, Zhang, Wang, Gao, Liu, Huang, and Wang]{sun2023emu}
Quan Sun, Qiying Yu, Yufeng Cui, Fan Zhang, Xiaosong Zhang, Yueze Wang, Hongcheng Gao, Jingjing Liu, Tiejun Huang, and Xinlong Wang.
\newblock Emu: Generative pretraining in multimodality.
\newblock \emph{arXiv preprint arXiv:2307.05222}, 2023{\natexlab{b}}.

\bibitem[Team(2024)]{team2024chameleon}
Chameleon Team.
\newblock Chameleon: Mixed-modal early-fusion foundation models.
\newblock \emph{arXiv preprint arXiv:2405.09818}, 2024.

\bibitem[Wang et~al.(2024{\natexlab{a}})Wang, Lu, Yang, Huang, Han, Hou, Zhang, and Xu]{wang2024illume}
Chunwei Wang, Guansong Lu, Junwei Yang, Runhui Huang, Jianhua Han, Lu~Hou, Wei Zhang, and Hang Xu.
\newblock Illume: Illuminating your llms to see, draw, and self-enhance.
\newblock \emph{arXiv preprint arXiv:2412.06673}, 2024{\natexlab{a}}.

\bibitem[Wang et~al.(2024{\natexlab{b}})Wang, Bai, Tan, Wang, Fan, Bai, Chen, Liu, Wang, Ge, et~al.]{wang2024qwen2}
Peng Wang, Shuai Bai, Sinan Tan, Shijie Wang, Zhihao Fan, Jinze Bai, Keqin Chen, Xuejing Liu, Jialin Wang, Wenbin Ge, et~al.
\newblock Qwen2-vl: Enhancing vision-language model's perception of the world at any resolution.
\newblock \emph{arXiv preprint arXiv:2409.12191}, 2024{\natexlab{b}}.

\bibitem[Wu et~al.(2025{\natexlab{a}})Wu, Chen, Wu, Ma, Liu, Pan, Liu, Xie, Yu, Ruan, et~al.]{wu2025janus}
Chengyue Wu, Xiaokang Chen, Zhiyu Wu, Yiyang Ma, Xingchao Liu, Zizheng Pan, Wen Liu, Zhenda Xie, Xingkai Yu, Chong Ruan, et~al.
\newblock Janus: Decoupling visual encoding for unified multimodal understanding and generation.
\newblock In \emph{Proceedings of the Computer Vision and Pattern Recognition Conference}, pages 12966--12977, 2025{\natexlab{a}}.

\bibitem[Wu et~al.(2025{\natexlab{b}})Wu, Gao, Ye, Li, Li, Guo, Liu, Xue, Hou, Liu, et~al.]{wu2025rewarddance}
Jie Wu, Yu~Gao, Zilyu Ye, Ming Li, Liang Li, Hanzhong Guo, Jie Liu, Zeyue Xue, Xiaoxia Hou, Wei Liu, et~al.
\newblock Rewarddance: Reward scaling in visual generation.
\newblock \emph{arXiv preprint arXiv:2509.08826}, 2025{\natexlab{b}}.

\bibitem[Xiao et~al.(2025)Xiao, Wang, Zhou, Yuan, Xing, Yan, Li, Wang, Huang, and Liu]{xiao2025omnigen}
Shitao Xiao, Yueze Wang, Junjie Zhou, Huaying Yuan, Xingrun Xing, Ruiran Yan, Chaofan Li, Shuting Wang, Tiejun Huang, and Zheng Liu.
\newblock Omnigen: Unified image generation.
\newblock In \emph{Proceedings of the Computer Vision and Pattern Recognition Conference}, pages 13294--13304, 2025.

\bibitem[Xie et~al.(2024)Xie, Mao, Bai, Zhang, Wang, Lin, Gu, Chen, Yang, and Shou]{xie2024show}
Jinheng Xie, Weijia Mao, Zechen Bai, David~Junhao Zhang, Weihao Wang, Kevin~Qinghong Lin, Yuchao Gu, Zhijie Chen, Zhenheng Yang, and Mike~Zheng Shou.
\newblock Show-o: One single transformer to unify multimodal understanding and generation.
\newblock \emph{arXiv preprint arXiv:2408.12528}, 2024.

\bibitem[Xu et~al.(2024)Xu, Liu, Wu, Tong, Li, Ding, Tang, and Dong]{xu2024imagereward}
Jiazheng Xu, Xiao Liu, Yuchen Wu, Yuxuan Tong, Qinkai Li, Ming Ding, Jie Tang, and Yuxiao Dong.
\newblock Imagereward: Learning and evaluating human preferences for text-to-image generation.
\newblock \emph{Advances in Neural Information Processing Systems}, 36, 2024.

\bibitem[Yang et~al.(2025)Yang, Li, Yang, Zhang, Hui, Zheng, Yu, Gao, Huang, Lv, et~al.]{yang2025qwen3}
An~Yang, Anfeng Li, Baosong Yang, Beichen Zhang, Binyuan Hui, Bo~Zheng, Bowen Yu, Chang Gao, Chengen Huang, Chenxu Lv, et~al.
\newblock Qwen3 technical report.
\newblock \emph{arXiv preprint arXiv:2505.09388}, 2025.

\bibitem[Yin et~al.(2023)Yin, Gharbi, Zhang, Shechtman, Durand, Freeman, and Park]{yin2023one}
Tianwei Yin, Micha{\"e}l Gharbi, Richard Zhang, Eli Shechtman, Fredo Durand, William~T Freeman, and Taesung Park.
\newblock One-step diffusion with distribution matching distillation.
\newblock \emph{arXiv preprint arXiv:2311.18828}, 2023.

\bibitem[Yin et~al.(2024)Yin, Gharbi, Park, Zhang, Shechtman, Durand, and Freeman]{yin2024improved}
Tianwei Yin, Micha{\"e}l Gharbi, Taesung Park, Richard Zhang, Eli Shechtman, Fredo Durand, and Bill Freeman.
\newblock Improved distribution matching distillation for fast image synthesis.
\newblock \emph{Advances in neural information processing systems}, 37:\penalty0 47455--47487, 2024.

\bibitem[Yu et~al.(2025)Yu, Chow, Yue, Pan, Wu, Wan, Li, Tang, Zhang, and Zhuang]{yu2025anyedit}
Qifan Yu, Wei Chow, Zhongqi Yue, Kaihang Pan, Yang Wu, Xiaoyang Wan, Juncheng Li, Siliang Tang, Hanwang Zhang, and Yueting Zhuang.
\newblock Anyedit: Mastering unified high-quality image editing for any idea.
\newblock In \emph{Proceedings of the Computer Vision and Pattern Recognition Conference}, pages 26125--26135, 2025.

\bibitem[Zhang et~al.(2024)Zhang, Wu, Ren, Xia, Kuang, Xie, Li, Xiao, Huang, Wen, et~al.]{zhang2024unifl}
Jiacheng Zhang, Jie Wu, Yuxi Ren, Xin Xia, Huafeng Kuang, Pan Xie, Jiashi Li, Xuefeng Xiao, Weilin Huang, Shilei Wen, et~al.
\newblock Unifl: Improve latent diffusion model via unified feedback learning.
\newblock \emph{Advances in Neural Information Processing Systems}, 37:\penalty0 67355--67382, 2024.

\bibitem[Zhang et~al.(2023)Zhang, Mo, Chen, Sun, and Su]{zhang2023magicbrush}
Kai Zhang, Lingbo Mo, Wenhu Chen, Huan Sun, and Yu~Su.
\newblock Magicbrush: A manually annotated dataset for instruction-guided image editing.
\newblock \emph{Advances in Neural Information Processing Systems}, 36:\penalty0 31428--31449, 2023.

\bibitem[Zheng et~al.(2023)Zheng, Chiang, Sheng, Zhuang, Wu, Zhuang, Lin, Li, Li, Xing, et~al.]{zheng2023judging}
Lianmin Zheng, Wei-Lin Chiang, Ying Sheng, Siyuan Zhuang, Zhanghao Wu, Yonghao Zhuang, Zi~Lin, Zhuohan Li, Dacheng Li, Eric Xing, et~al.
\newblock Judging llm-as-a-judge with mt-bench and chatbot arena.
\newblock \emph{Advances in neural information processing systems}, 36:\penalty0 46595--46623, 2023.

\bibitem[Zheng et~al.(2024)Zheng, Yin, Xie, Sun, Huang, Yu, Cao, Kozyrakis, Stoica, Gonzalez, et~al.]{zheng2024sglang}
Lianmin Zheng, Liangsheng Yin, Zhiqiang Xie, Chuyue~Livia Sun, Jeff Huang, Cody~Hao Yu, Shiyi Cao, Christos Kozyrakis, Ion Stoica, Joseph~E Gonzalez, et~al.
\newblock Sglang: Efficient execution of structured language model programs.
\newblock \emph{Advances in neural information processing systems}, 37:\penalty0 62557--62583, 2024.

\bibitem[Zhou et~al.(2024)Zhou, Yu, Babu, Tirumala, Yasunaga, Shamis, Kahn, Ma, Zettlemoyer, and Levy]{zhou2024transfusion}
Chunting Zhou, Lili Yu, Arun Babu, Kushal Tirumala, Michihiro Yasunaga, Leonid Shamis, Jacob Kahn, Xuezhe Ma, Luke Zettlemoyer, and Omer Levy.
\newblock Transfusion: Predict the next token and diffuse images with one multi-modal model.
\newblock \emph{arXiv preprint arXiv:2408.11039}, 2024.

\end{thebibliography}

\clearpage



\end{document}